\documentclass{article}

    \PassOptionsToPackage{numbers, sort&compress}{natbib}
 \usepackage[preprint]{neurips_2026}

\usepackage[utf8]{inputenc} 
\usepackage[T1]{fontenc}    
\usepackage{hyperref}       
\usepackage{url}            
\usepackage{booktabs}       
\usepackage{amsfonts}       
\usepackage{nicefrac}       
\usepackage{microtype}      
\usepackage{xcolor}         

\usepackage{amsmath,amssymb}
\usepackage{algorithm,algorithmic}
\usepackage{graphicx}
\usepackage{multirow,makecell}
\usepackage{wrapfig}

\title{GRACE: Gradient-aligned Reasoning Data Curation for Efficient Post-training}

\author{
Junjie Li\thanks{These authors contributed equally to this work.} \\
Harbin Institute of Technology, Shenzhen, China \\
\texttt{22b351018@stu.hit.edu.cn} \\
\And
Ziao Wang\footnotemark[1] \\
Hong Kong Baptist University, China \\
\texttt{ziaowang@hkbu.edu.cn} \\
\And
NingXuan Ma \\
Harbin Institute of Technology, Shenzhen, China \\
\texttt{2023311G27@stu.hit.edu.cn} \\
\And
Jianghong Ma\thanks{Corresponding author.} \\
\small Harbin Institute of Technology, Shenzhen, China \\
\small City University of Hong Kong, China \\
\texttt{majianghong@hit.edu.cn} \\
\And
Xiaofeng Zhang\footnotemark[2] \\
Harbin Institute of Technology, Shenzhen, China \\
\texttt{zhangxiaofeng@hit.edu.cn} \\
}

\begin{document}

\maketitle

\begin{abstract}
  Existing reasoning data curation pipelines score whole samples, treating every intermediate step as equally valuable. In reality, steps within a trace contribute very unevenly, and selecting reasoning data well requires assessing them individually. We present GRACE, a gradient-aligned curation method that views each reasoning trace as a sequence of optimization events and scores every step by two complementary signals: its alignment with the answer-oriented gradient direction, and its consistency with the preceding reasoning trajectory. Step-level scores are aggregated into a sample-level value for subset selection, using only the model's internal optimization signals and no external reward models or step annotations. To make this scalable, GRACE introduces a representation-level gradient proxy that estimates step-level alignment from token-level upstream signals in a single forward pass. Post-training Qwen3-VL-2B-Instruct on MMathCoT-1M, GRACE reaches 108.8\% of the full-data performance with 20\% of the data and retains 100.2\% with only 5\%, with subsets that transfer effectively across model backbones.
\end{abstract}

\section{Introduction}

Large-scale reasoning datasets have become a cornerstone for post-training large language and vision-language models~\citep{wei2022chain,xu2025llavacot}. The standard way to use them is to supervise the model on the entire reasoning trace, treating every step as an equally valuable target. In reality, the steps within a trace contribute very unevenly: some directly support the final answer, while others restate earlier content, explore irrelevant tangents, or introduce noise. Training on them uniformly wastes budget on low-value steps and dilutes the contribution of useful ones. This cost has become significant as reasoning corpora grow to millions of traces~\citep{mmathcot} and post-training takes hundreds of GPU-hours per run. Choosing what to train on therefore matters~\citep{lima}, and for reasoning data this means assessing steps individually rather than ranking whole traces.



Existing data curation methods improve training efficiency by selecting samples based on correctness~\citep{wang2023selfconsistency}, reward models~\citep{ouyang2022training}, or sample-level influence~\citep{icons,tive,coincide} and all of them operate at the granularity of entire traces.
As a result, a trace with correct final answers but poor intermediate steps~\citep{uesato2022solving} is treated as equally valuable as a tightly reasoned one. This highlights a fundamental limitation: \textit{current approaches lack a mechanism to assess how individual reasoning steps contribute to optimization}.

\begin{figure}
  \centering
  \includegraphics[width=1.0\linewidth]{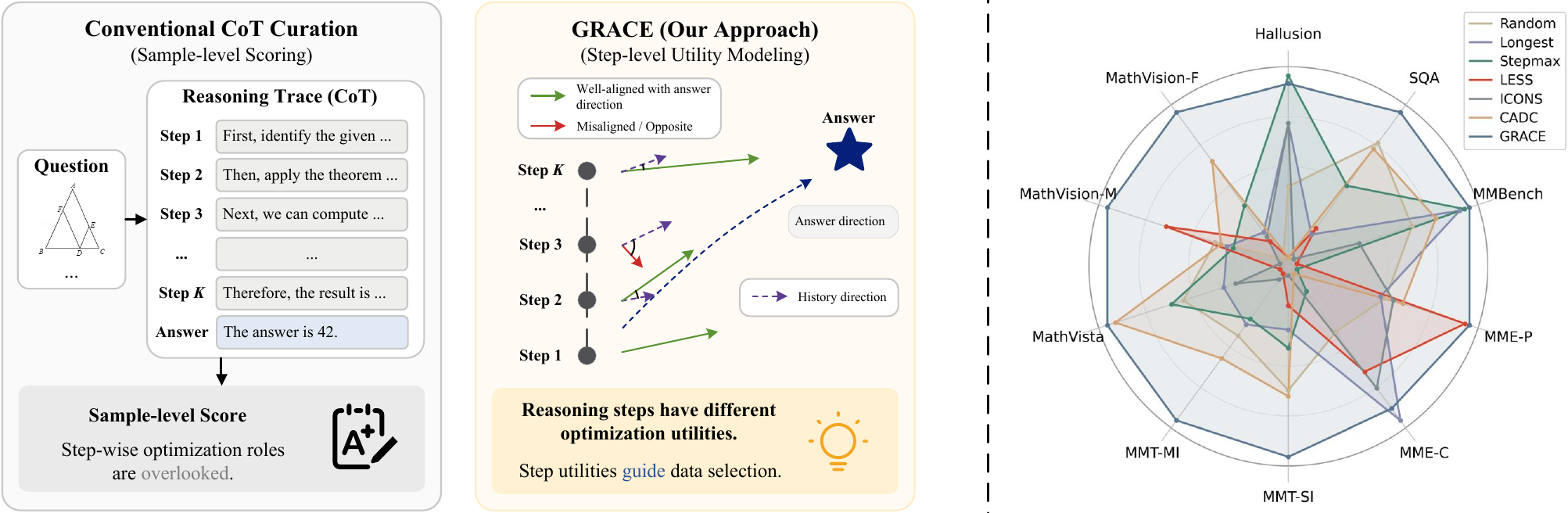}
  \caption{Motivation and empirical effect of GRACE.
    Left: Reasoning traces are viewed as sequences of optimization events, where each step induces an update direction whose utility depends on its alignment with the target objective and the evolving trajectory.
    Right: Radar chart comparing downstream performance across benchmarks. GRACE achieves full-data or better performance using only a fraction of the training data.} 
  \label{fig:motivation}
\end{figure}

In this work, we revisit reasoning data from an optimization perspective. Instead of viewing reasoning traces as static supervision targets, we model them as sequences of optimization events, where each reasoning step induces a local training signal that affects the gradient direction toward the final answer. From this view, the utility of reasoning data depends not only on external attributes such as correctness or length, but also on whether its intermediate steps constructively support optimization.

Building on this view, we propose \textbf{GRACE} (\textbf{G}radient-aligned \textbf{R}easoning d\textbf{A}ta \textbf{C}uration for \textbf{E}fficient post-training), a method that performs fine-grained data curation by estimating step-level optimization utility. Rather than pruning or rewriting reasoning traces, GRACE assigns each step a utility score based on two complementary criteria: (i) its alignment with the answer-oriented optimization direction, and (ii) its consistency with the accumulated reasoning trajectory. These signals capture both task-driven and trajectory-aware contributions of each step. The resulting step-level scores are then aggregated to produce a sample-level utility score, enabling effective subset selection while preserving the simplicity of sample-level training.
Fig.~\ref{fig:motivation} illustrates the motivation of GRACE and provides empirical evidence that optimization-aware curation can retain strong performance with substantially fewer training samples.

A key challenge is that true step-level gradients are computationally intractable at scale. Naively computing gradients for each reasoning step would require decomposing traces into multiple training instances and performing repeated backward passes. To address this, GRACE introduces a \textbf{representation-level gradient proxy} that approximates step-induced optimization directions using token-level upstream signals. This proxy enables efficient estimation of step-level alignment from a single forward pass, making optimization-aware curation practical for large-scale CoT datasets.


We evaluate GRACE by post-training Qwen3-VL-2B-Instruct~\citep{qwen3vl} on MMathCoT-1M~\citep{mmathcot} and assessing the resulting models on a diverse suite of multimodal benchmarks spanning mathematical reasoning and general visual question answering. GRACE consistently identifies high-value subsets: training on only \textbf{20\%} of the curated data surpasses full-data performance, reaching \textbf{108.8\%} of the full-data result averaged across benchmarks, while using only \textbf{5\%} retains \textbf{100.2\%}. Furthermore, the selected subsets transfer effectively across model backbones, suggesting that the proposed optimization-based signal captures intrinsic data value beyond a specific model configuration.

Our contributions are three-fold:
\begin{enumerate}
    \item We introduce an optimization-based perspective on reasoning data, framing reasoning traces as sequences of optimization events and highlighting the role of step-level alignment in effective learning.
    \item We propose GRACE, a reasoning data curation method that aggregates step-level optimization signals derived from answer-oriented alignment and trajectory consistency for sample-level subset selection.
    \item We develop a representation-level gradient proxy that enables scalable estimation of step-level alignment without per-step parameter-space gradient computation.
\end{enumerate}

\section{Method}
\label{sec:method}

\begin{figure*}[t]
    \centering
    \includegraphics[width=1.0\textwidth]{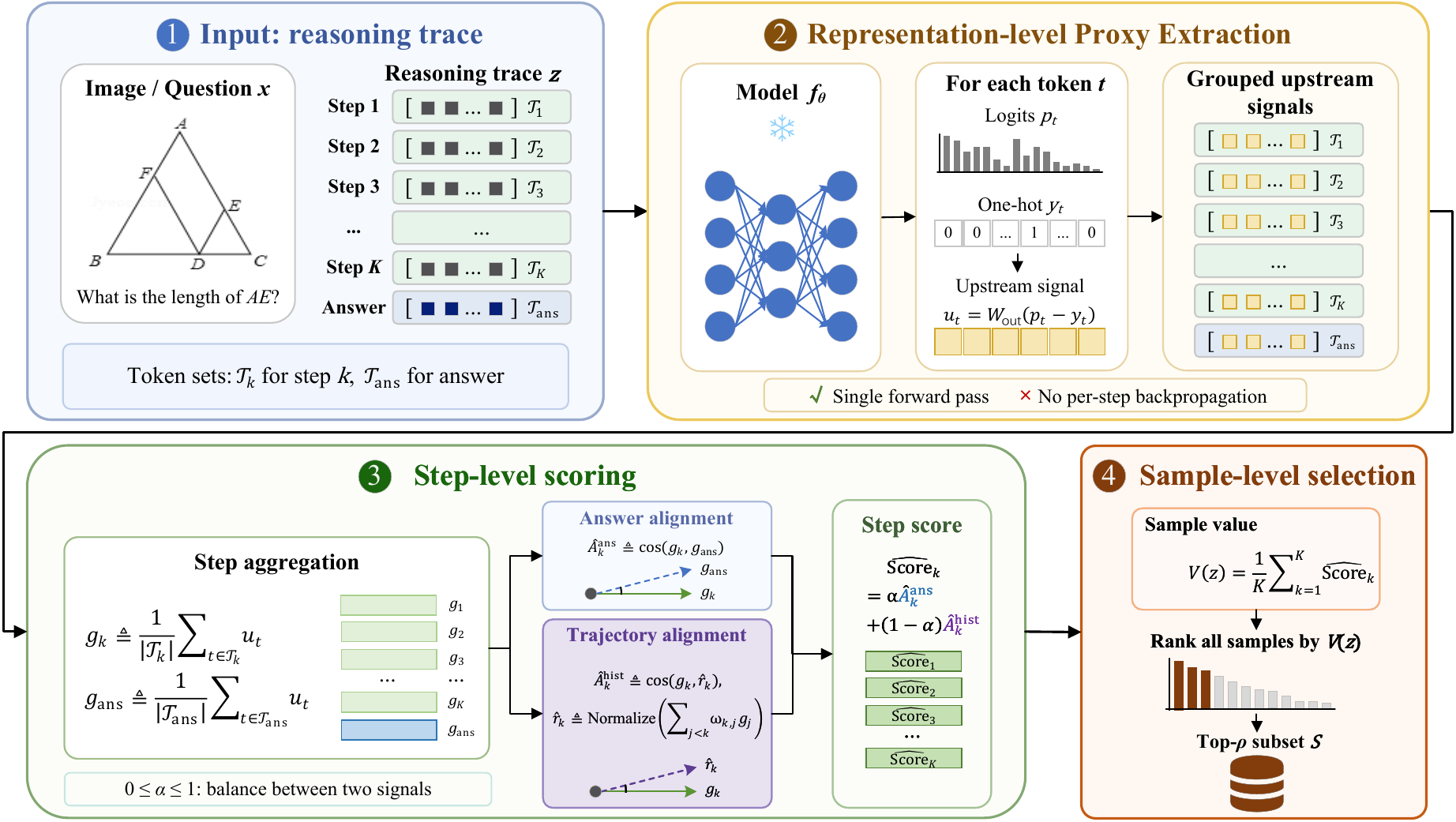}
    \caption{
    The GRACE curation pipeline. \textbf{(1)} Given an input and its reasoning trace, GRACE identifies token sets for each step and the answer. \textbf{(2)} A fixed scoring model extracts token-level upstream signals in one forward pass and groups them by token sets. \textbf{(3)} Grouped signals are averaged into gradient proxies and scored by answer and trajectory alignment. \textbf{(4)} Step scores are aggregated into a sample value for ranking and top-$\rho$ subset selection.
    }
    \label{fig:overview}
\end{figure*}

In this section, we present GRACE. We first define the reasoning data curation problem, then derive the step-level optimization utility, introduce its scalable representation-level proxy, and describe sample-level subset selection.
The overall pipeline is illustrated in Fig.~\ref{fig:overview}.

\subsection{Problem Formulation}
\label{sec:problem}

We consider a reasoning dataset $\mathcal{D}=\{z_i\}_{i=1}^{N}$, where each sample is
\[
z_i=(x_i,\mathbf{s}_i,a_i),
\]
with $x_i$ denoting the input, $\mathbf{s}_i=(s_{i,1},s_{i,2},\dots,s_{i,K_i})$ denoting a sequence of reasoning steps, and $a_i$ denoting the final answer. 
Let $f_\theta$ denote the model, and let $\mathcal{T}_{i,k}$ and $\mathcal{T}_{i,\mathrm{ans}}$ denote the token positions of step $s_{i,k}$ and the answer segment, respectively.
For any token set $\mathcal{T}$, we define the average token-level loss
\[
L(\theta;z_i,\mathcal{T})
=
\frac{1}{|\mathcal{T}|}
\sum_{t\in\mathcal{T}} L_t(\theta;z_i).
\]
Accordingly, the step loss and answer loss are given by
$L_{i,k}=L(\theta;z_i,\mathcal{T}_{i,k})$ and 
$L_i^{\mathrm{ans}}=L(\theta;z_i,\mathcal{T}_{i,\mathrm{ans}})$, 
and the full loss over the reasoning trace and answer is
\[
\mathcal{T}_{i,\mathrm{full}}
=
\bigcup_{k=1}^{K_i}\mathcal{T}_{i,k}\cup \mathcal{T}_{i,\mathrm{ans}},
\qquad
L_{\mathrm{full}}(\theta;z_i)=L(\theta;z_i,\mathcal{T}_{i,\mathrm{full}}).
\]
Standard post-training minimizes
\begin{equation}
\mathcal{L}_{\mathrm{SFT}}(\theta;\mathcal{D})
=
\frac{1}{|\mathcal{D}|}
\sum_{z_i\in\mathcal{D}}
L_{\mathrm{full}}(\theta;z_i).
\end{equation}

Our goal is to select a compact subset $\mathcal{S}\subset \mathcal{D}$ with budget $|\mathcal{S}|=\lceil \rho|\mathcal{D}| \rceil$, where $\rho\in(0,1)$ is the selection ratio, such that post-training on $\mathcal{S}$ preserves or improves downstream performance compared with training on the full dataset. To this end, GRACE assigns each sample a scalar value score $V(z_i)$ and selects the top-ranked subset:
\begin{equation}
   \mathcal{S}
=
\left\{
z_i\in\mathcal{D}\mid
\operatorname{rank}_{V}(z_i)
\le
\lceil \rho|\mathcal{D}|\rceil
\right\}.
\label{eq:subset_problem}
\end{equation}


The key question is how to define $V(z_i)$ for reasoning data. 
GRACE addresses this by estimating the optimization utility of each reasoning step and aggregating these step-level signals into a sample-level value.
Since these values are computed from model-internal signals, the scoring model should provide stable representations.
Following prior works~\citep{less,cadc}, we obtain the scoring model $f_\theta$ by warming up an initial model $f_{\theta_0}$ on a $\gamma$-ratio subset of $\mathcal{D}$, and keep it fixed during data scoring.

\subsection{Step-level Optimization Utility}
\label{sec:step}


We define the utility of a reasoning step based on its contribution to optimizing a target objective. For clarity, we omit the sample index $i$ when discussing a single sample $z=(x,\mathbf{s},a)$. Let $\mathcal{T}_k$ denote the token set of step $s_k$, and write
$L_k(\theta;z)=L(\theta;z,\mathcal{T}_k)$.
We adopt a standard first-order influence perspective~\citep{tracin,koh2017influence}. 
Consider a small update induced by step $s_k$:
\begin{equation}
\theta' = \theta - \eta \nabla_\theta L_k(\theta;z),
\label{eq:step_update}
\end{equation}
where $\eta>0$ is the learning rate. 
Let $L^{\mathrm{tar}}(\theta;z)$ denote a target loss specifying the desired optimization direction; under Eq.~\ref{eq:step_update}, its first-order change is approximated as
\begin{equation}
\begin{aligned}
L^{\mathrm{tar}}(\theta';z)-L^{\mathrm{tar}}(\theta;z)
&\approx
-\eta
\left\langle
\nabla_\theta L_k,
\nabla_\theta L^{\mathrm{tar}}
\right\rangle \\
&=
-\eta
\|\nabla_\theta L_k\|
\,
\|\nabla_\theta L^{\mathrm{tar}}\|
\,
\cos\bigl(
\nabla_\theta L_k,
\nabla_\theta L^{\mathrm{tar}}
\bigr).
\end{aligned}
\label{eq:first_order_target}
\end{equation}

This shows that a step is locally beneficial when its induced gradient is directionally aligned with the target gradient. 
Since step lengths and gradient scales can vary substantially across reasoning segments, we focus on the normalized directional component:
\begin{equation}
A_k^{\mathrm{tar}}
\triangleq
\cos\bigl(
\nabla_\theta L_k,
\nabla_\theta L^{\mathrm{tar}}
\bigr).
\label{eq:tar_direction}
\end{equation}
See Appendix~\ref{app:directional_step_utility} for details.
Different choices of target direction correspond to different notions of step utility. 
In GRACE, we consider two complementary objectives:

\textbf{(1) Answer-oriented objective.}  
We instantiate $L^{\mathrm{tar}} = L^{\mathrm{ans}}$, where
$L^{\mathrm{ans}}(\theta;z)=L(\theta;z,\mathcal{T}_{\mathrm{ans}})$
is the loss on the answer segment and $\mathcal{T}_{\mathrm{ans}}$ denotes its token set. This gives
\begin{equation}
    A_k^{\mathrm{ans}}
\triangleq
\cos\bigl(
\nabla_\theta L_k,
\nabla_\theta L^{\mathrm{ans}}
\bigr).
\label{eq:ans_direction}
\end{equation}
which measures whether the step supports optimizing the final answer. While the answer-oriented objective captures whether a step supports the final answer, it does not characterize whether the step is coherent with the reasoning process that precedes it.

\textbf{(2) Trajectory consistency objective.}  
Reasoning steps form an ordered trajectory rather than independent supervision signals. 
For steps with preceding context, we define the historical reference direction and its corresponding alignment score jointly:
\begin{equation}
A_k^{\mathrm{hist}} \triangleq
\cos\bigl(
\nabla_\theta L_k,
r_k
\bigr),
\qquad
r_k
\triangleq
\mathrm{Normalize}
\left(
\sum_{j<k}\omega_{k,j}\nabla_\theta L_j
\right),
\label{eq:hist_direction}
\end{equation}

where $\omega_{k,j}\ge 0$ and $\sum_{j<k}\omega_{k,j}=1$ control the contribution of each previous step to the historical reference direction.
The score $A_k^{\mathrm{hist}}$ measures whether the current step continues the existing reasoning trajectory.

Combining these two criteria, the final step-level utility is defined as
\begin{equation}
\mathrm{Score}_k
\triangleq
\alpha A_k^{\mathrm{ans}}
+
(1-\alpha)A_k^{\mathrm{hist}},
\qquad \alpha \in [0,1],
\label{eq:step_score}
\end{equation}
where the historical term is omitted for $k=1$, since no preceding context exists.

\subsection{Representation-level Gradient Proxy}
\label{sec:proxy}

The step utility in Eq.~\ref{eq:step_score} is defined through parameter-space gradients. Directly evaluating such gradients for every reasoning step would require isolating each step as a separate loss and performing repeated backward passes over decomposed traces. To obtain a scalable estimate of step-level alignment, we project the optimization signal to the final representation interface and compute alignment in this lower-dimensional space.

Since the full gradient can be decomposed as $\nabla_{\theta} L
=
\left[
\nabla_{W_{\mathrm{out}}} L ;
\nabla_{\theta_{\mathrm{rep}}} L
\right]$, we estimate step-induced directions in the representation-producing subspace $\theta_{\mathrm{rep}}$, where all reasoning tokens interact through the model's internal features.
Let $h_t\in\mathbb{R}^d$ denote the final-layer hidden state at token position $t$, where $d$ is the hidden dimension, and let $W_{\mathrm{out}}\in \mathbb{R}^{d\times V}$ be the output projection matrix, where $V$ is the vocabulary size. The pre-softmax logit is $\ell_t=W_{\mathrm{out}}^\top h_t\in\mathbb{R}^{V}$, and the next-token probability is $p_t=\mathrm{softmax}(\ell_t)$. For the token-level cross-entropy loss $L_t$,
\[
\frac{\partial L_t}{\partial \ell_t}=p_t-y_t,
\]
where $y_t$ is the one-hot target token. The corresponding gradient at the representation interface is
\begin{equation}
u_t
\triangleq
\frac{\partial L_t}{\partial h_t}
=
W_{\mathrm{out}}(p_t-y_t).
\label{eq:upstream_gradient}
\end{equation}

Let $\theta_{\mathrm{rep}}$ denote the parameters that produce the final-layer representations, and define $J_t=\partial h_t/\partial \theta_{\mathrm{rep}}$. By the chain rule, the representation-parameter gradient induced by token $t$ is
$
\nabla_{\theta_{\mathrm{rep}}} L_t
=
J_t^\top u_t .
$
Accordingly, for a token set $\mathcal{T}$, the corresponding segment gradient is
$
\nabla_{\theta_{\mathrm{rep}}} L(\mathcal{T})
=
|\mathcal{T}|^{-1}
\sum_{t\in\mathcal{T}}
J_t^\top u_t .
$
Thus, $\{u_t\}$ are the common upstream optimization signals that drive updates of the representation-producing parameters through the Jacobian mapping.

Eq.~\ref{eq:tar_direction} requires the cosine alignment between the update directions induced by two token segments. For two token sets $\mathcal{T}_1$ and $\mathcal{T}_2$, its representation-parameter form is
\begin{equation}
\cos\!\left(
\nabla_{\theta_{\mathrm{rep}}} L(\mathcal{T}_1),
\nabla_{\theta_{\mathrm{rep}}} L(\mathcal{T}_2)
\right) 
=
\frac{
\sum_{t\in\mathcal{T}_1}
\sum_{t'\in\mathcal{T}_2}
u_t^\top
\left(J_tJ_{t'}^\top\right)
u_{t'}
}{
\left\|
\sum_{t\in\mathcal{T}_1} J_t^\top u_t
\right\|
\left\|
\sum_{t'\in\mathcal{T}_2} J_{t'}^\top u_{t'}
\right\|
}.
\label{eq:jacobian_cosine}
\end{equation}

Eq.~\ref{eq:jacobian_cosine} shows that exact representation-parameter alignment depends on both token-level upstream signals and Jacobian-induced interactions $J_tJ_{t'}^\top$. 

Exact step-level evaluation of these interactions would require isolating each step loss and backpropagating it through $\theta_{\mathrm{rep}}$, leading to repeated per-step gradient computation. 
To make step-level scoring scalable, we introduce an interface-level surrogate that preserves upstream optimization signals while avoiding explicit construction of the Jacobian-induced geometry. 
This design follows scalable data valuation methods that approximate gradient information in proxy spaces~\citep{less,badge,trak}. 
Under this surrogate, the segment-level update direction is represented by the aggregated upstream signal:
\begin{equation}
g(\mathcal{T})
\triangleq
\frac{1}{|\mathcal{T}|}
\sum_{t\in\mathcal{T}}u_t.
\label{eq:proxy_general}
\end{equation}
The averaging follows the definition of the segment loss $L(\theta;z,\mathcal{T})$ as a token-level mean, ensuring that proxy directions are not biased by segment length.

Using the token sets $\mathcal{T}_k$ and $\mathcal{T}_{\mathrm{ans}}$, we write $g_k\triangleq g(\mathcal{T}_k)$ and $g_{\mathrm{ans}}\triangleq g(\mathcal{T}_{\mathrm{ans}})$. The step-level utility is then computed in the proxy space as
\begin{equation}
\widehat{\mathrm{Score}}_k
=
\begin{cases}
\widehat{A}_k^{\mathrm{ans}}, & k=1,\\[2mm]
\alpha \widehat{A}_k^{\mathrm{ans}}
+
(1-\alpha)\widehat{A}_k^{\mathrm{hist}}, & k>1,
\end{cases}
\label{eq:proxy_score}
\end{equation}
where
\[
\widehat{A}_k^{\mathrm{ans}}
\triangleq
\cos(g_k,g_{\mathrm{ans}}),
\qquad
\widehat{A}_k^{\mathrm{hist}}
\triangleq
\cos(g_k,\widehat{r}_k),
\qquad
\widehat{r}_k
\triangleq
\mathrm{Normalize}\left(\sum_{j<k}\omega_{k,j}g_j\right).
\]
Importantly, for each sample, the fixed output projection $W_{\mathrm{out}}$, the forward-pass probabilities $\{p_t\}$, and the ground-truth tokens $\{y_t\}$ are sufficient to obtain directional gradient proxies for all reasoning steps in a single forward pass, without constructing per-step training instances or performing backward passes.
Detailed derivation is provided in Appendix~\ref{app:proxy_derivation}.

\subsection{Sample-level Aggregation}
\label{sec:aggregation}

Given the step-level proxy utility $\widehat{\mathrm{Score}}_{i,k}$ in Eq.~\ref{eq:proxy_score}, 
GRACE instantiates the sample value $V(z_i)$ by averaging step-level utilities:
\begin{equation}
V(z_i)
=
\frac{1}{K_i}
\sum_{k=1}^{K_i}
\widehat{\mathrm{Score}}_{i,k}.
\label{eq:aggregation}
\end{equation}
This aggregation treats the reasoning trace as a sequence of optimization events and measures its overall training value by the average step-level utility.


The samples are ranked by $V(z_i)$ and selected according to the top-budget rule in Eq.~\ref{eq:subset_problem}. 
The selected subset $\mathcal{S}$ is used for post-training with the original reasoning traces. 
See Appendix~\ref{app:algorithm} for details.

\section{Experiments}
\label{sec:experiments}

We empirically validate GRACE through a series of experiments designed to answer five questions:
\begin{itemize}
    \item Does GRACE curate reasoning data more effectively than existing methods (Sec.~\ref{sec:exp_main})?
    \item Do the curated subsets transfer across model backbones (Sec.~\ref{sec:exp_transfer})?
    \item Which components of GRACE drive its effectiveness (Sec.~\ref{sec:exp_ablation})?
    \item How robust is GRACE to its design hyperparameters (Sec.~\ref{sec:exp_hyper})?
    \item What is the computational cost of GRACE relative to gradient-based alternatives (Sec.~\ref{sec:exp_cost})?
\end{itemize}


\paragraph{Experimental Setup.}
We evaluate GRACE by post-training Qwen3-VL-2B-Instruct on the reasoning-rich candidate pool of MMathCoT-1M and comparing against heuristic selectors (Random, Longest, Stepmax) and data-curation baselines (LESS, ICONS, CADC) under the same selection budget and training recipe.
Evaluation covers general VQA/perception, multi-task and multi-image reasoning, and mathematical reasoning benchmarks, with \textit{Rel. Avg.} denoting performance normalized by the full-data baseline.
Unless otherwise stated, GRACE uses $\rho=0.2$, $\gamma=0.05$, uniform history aggregation, and $\alpha=0.7$.
Full experimental details, including datasets, backbones, benchmarks, training recipes, and hardware, are provided in Appendix~\ref{app:experimental_details}.

\begin{table}[tb]
\centering
  \caption{Performance of Qwen3-VL-2B under the data selection methods.
    \textit{Data \%} denotes the proportion of training data used, and \textit{Rel. Avg.} is the average relative performance over benchmarks.
    \textuparrow~indicates larger is better.
    \textbf{Bold} and \underline{underlined} values denote the best and second-best results among 20\% data selection methods, respectively.}
  \label{tab:grace_performance}
\resizebox{\linewidth}{!}{
\begin{tabular}{llccccccccccc}
\toprule
\multirow{2}[2]{*}{\textbf{Method}} 
& \multirow{2}[2]{*}{\textbf{Data \%}} 
& \multirow{2}[2]{*}{\textbf{Hallusion\citep{hallusionbench}\textuparrow}} 
& \multirow{2}[2]{*}{\textbf{SQA\citep{scienceqa}\textuparrow}} 
& \multirow{2}[2]{*}{\textbf{MMBench\citep{mmbench}\textuparrow}} 
& \multicolumn{2}{c}{\textbf{MME\citep{mme}\textuparrow}} 
& \multicolumn{2}{c}{\textbf{MMT\citep{mmtbench}\textuparrow}} 
& \multirow{2}[2]{*}{\textbf{MathVista\citep{mathvista}\textuparrow}} 
& \multicolumn{2}{c}{\textbf{MathVision\citep{mathvision}\textuparrow}} 
& \multirow{2}[2]{*}{\textbf{Rel. Avg.\textuparrow}} \\
\cmidrule(lr){6-7}\cmidrule(lr){8-9}\cmidrule(lr){11-12}
& & & & 
& \textbf{Perc.} & \textbf{Cog.} 
& \textbf{SI} & \textbf{MI} 
& 
& \textbf{MINI} & \textbf{Full} 
& \\
\midrule
\textbf{Full} & 100\% & 43.7 & 80.2 & 69.3 & 1517.3 & 656.1 & 55.0 & 53.4 & 52.5 & 16.4 & 14.7 & -- \\
\midrule
\textbf{Random} & 20\% & 45.5 & \underline{84.5} & 72.6 & 1495.5 & 653.2 & 57.4 & 54.8 & 52.3 & 16.4 & 13.9 & 101.4 \\
\textbf{Longest} & 20\% & 46.3 & 83.0 & 73.6 & 1494.5 & \textbf{687.5} & 56.4 & 54.6 & 51.3 & 15.8 & 14.5 & 101.7 \\
\textbf{Stepmax} & 20\% & \textbf{46.9} & 83.8 & \underline{73.7} & 1477.7 & 637.5 & 56.7 & 54.5 & 52.6 & 15.5 & 15.1 & 101.5 \\
\textbf{LESS~\cite{less}} & 20\% & 44.6 & 83.1 & 70.2 & \underline{1511.4} & 668.6 & 56.0 & 53.7 & 49.9 & \underline{18.8} & 14.3 & 101.7 \\
\textbf{ICONS~\cite{icons}} & 20\% & 46.3 & 82.6 & 71.5 & 1497.0 & 675.0 & 55.5 & 53.8 & 51.0 & 13.2 & 14.4 & 99.1 \\
\textbf{CADC~\cite{cadc}} & 20\% & 44.6 & 84.4 & 73.1 & 1498.9 & 630.7 & \underline{57.5} & \underline{55.2} & \underline{54.0} & 16.1 & \underline{16.1} & \underline{102.6} \\
\midrule
\multirow{4}{*}{\makecell[l]{\textbf{GRACE}\\\textbf{(Ours)}}}
& 5\%  & 48.5 & 83.0 & 71.1 & 1509.3 & 617.5 & 56.5 & 55.1 & 51.9 & 14.0 & 14.8 & 100.2 \\
& 10\% & 45.7 & 82.5 & 69.5 & 1500.7 & 660.4 & 57.7 & 55.1 & 53.0 & 18.1 & 15.5 & 103.2 \\
& 15\% & 48.5 & 82.6 & 71.1 & 1500.5 & 679.3 & 57.6 & 54.9 & 54.3 & 17.8 & 16.7 & 105.2 \\
& 20\% & \underline{46.8} & \textbf{85.0} & \textbf{73.8} & \textbf{1512.3} & \underline{682.9} & \textbf{58.5} & \textbf{56.3} & \textbf{54.2} & \textbf{21.7} & \textbf{17.2} & \textbf{108.8} \\
\bottomrule
\end{tabular}%
}
\end{table}

\subsection{Main Results}
\label{sec:exp_main}
 
Table~\ref{tab:grace_performance} reports per-benchmark performance on Qwen3-VL-2B, comparing all baselines at a $20\%$ selection ratio against GRACE at $5\%$, $10\%$, $15\%$, and $20\%$. GRACE consistently outperforms all heuristic and gradient-based baselines: at $20\%$ data, it reaches $108.8\%$ relative average, surpassing the strongest baseline (CADC, $102.6\%$) by a substantial margin and exceeding the full-data baseline by $8.8$ points. With only $5\%$ of the data, GRACE retains $100.2\%$ of full-data performance, demonstrating that step-level optimization-aware scoring can identify highly compact yet informative subsets.
This gain over full-data training is partly due to continued SFT on an instruction-tuned backbone: the math-centric candidate pool may over-specialize the model toward mathematical reasoning and cause partial forgetting on general VQA and perception abilities. 
Thus, reduced subsets can sometimes outperform full-data SFT, while GRACE further improves this effect by selecting traces with more favorable step-level optimization signals.
 
Table~\ref{tab:qwen3vl2b_rel_avg} compares relative average performance across $5\%$--$15\%$ selection ratios. GRACE is the only method that exceeds full-data performance at every ratio, while heuristic baselines fluctuate around the full-data line and gradient-based baselines are unstable at low ratios.

\begin{table}[tb]
\centering
\caption{Relative average performance (\%) of Qwen3-VL-2B under different data selection ratios.
\textit{Data \%} denotes the proportion of data used.
Values are normalized to the full-data training baseline.
}
\label{tab:qwen3vl2b_rel_avg}
\small
\setlength{\tabcolsep}{4pt}
\begin{tabular}{lcccccccc}
\toprule
\textbf{Data \%} 
& \textbf{Full} 
& \textbf{Random} 
& \textbf{Longest} 
& \textbf{Stepmax} 
& \textbf{LESS} 
& \textbf{ICONS} 
& \textbf{CADC} 
& \textbf{GRACE (Ours)} \\
\midrule
5\%  & 100.0 & 96.4  & 98.3  & 95.6  & 90.4 & 95.9  & \underline{99.8}  & \textbf{100.2} \\
10\% & 100.0 & 98.5  & 100.5 & 98.8  & 95.5 & \underline{102.4} & 101.3 & \textbf{103.2} \\
15\% & 100.0 & 100.0 & 98.2  & 100.6 & 98.9 & \underline{101.7} & 100.7 & \textbf{105.2} \\
\bottomrule
\end{tabular}
\end{table}

\subsection{Transfer Across Backbones}
\label{sec:exp_transfer}
 
To test whether the value assigned by GRACE reflects data properties that transfer beyond the scoring backbone, we post-train other backbones on the subset selected using Qwen3-VL-2B, \emph{without} re-running data selection. 
We consider Qwen2.5-VL-3B~\citep{qwen2.5-VL}, LLaVA-1.5-7B~\citep{llava15}, and Qwen3-VL-8B~\citep{qwen3vl}, covering different model families and scales, and compare them with their corresponding full-data baselines under the same training recipe.

\begin{wrapfigure}{r}{0.4\linewidth}  
    \centering      
    \resizebox{\linewidth}{!}{
        \includegraphics[width=\textwidth]{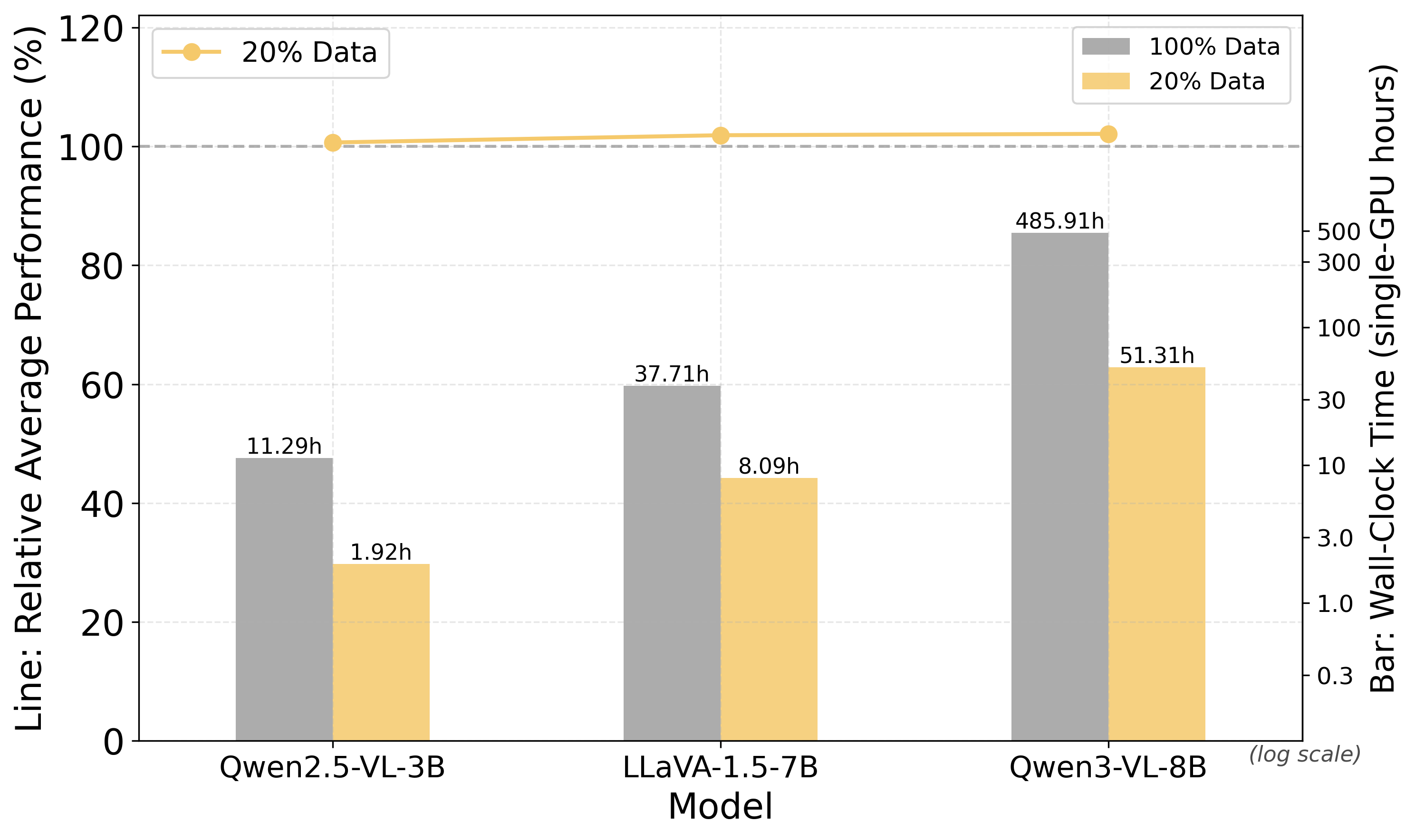}
        }
    \caption{
    Transferability and training efficiency across backbones.
    }
    \label{fig:transfer_efficiency}
\end{wrapfigure}
Figure~\ref{fig:transfer_efficiency} further examines whether GRACE-selected data transfer across backbones.
The line reports the relative average performance of the 20\% GRACE subset, normalized by each backbone's full-data baseline. 
The dashed horizontal line marks the 100\% full-data level, and the bars compare single-GPU wall-clock training time under 100\% and 20\% data.
We select the 20\% subset once using Qwen3-VL-2B and reuse it to post-train Qwen2.5-VL-3B, LLaVA-1.5-7B, and Qwen3-VL-8B without re-running data selection.
The curated subset consistently surpasses the corresponding full-data baseline across all backbones, indicating that the value captured by GRACE is not tied to a single scoring model.
This suggests that step-level optimization utility reflects transferable properties of reasoning data rather than model-specific artifacts.

In addition to transferability, the selected subset substantially reduces training cost, with larger savings on larger backbones where full-data post-training is more expensive.
Overall, GRACE improves the efficiency--performance trade-off: a compact reasoning subset can match or surpass full-data performance while requiring only a fraction of the training time.

\subsection{Ablation Studies}
\label{sec:exp_ablation}
 

We ablate the two utility components and gradient proxy on Qwen3-VL-2B at $\rho=0.2$. 
We consider five variants:
(a) \textbf{w/o historical score}: setting $\alpha=1$, scoring each step only by answer-oriented alignment;
(b) \textbf{w/o answer score}: removing the answer-oriented term and scoring steps only by historical alignment, with the first step omitted since no history exists;
(c) \textbf{Target = CoT + Ans}: using the union of all reasoning steps and the answer as the optimization target;
(d) \textbf{Target = Suffix}: using the trailing suffix after the current step, testing whether local future context suffices as an answer surrogate;
(e) \textbf{Proxy = projected gradient}: replacing $g(\mathcal{T})$ with LESS-style projected parameter-space gradients computed per step, while keeping the rest of GRACE unchanged.


\begin{table}[tb]
\centering
\caption{Ablation study of GRACE on Qwen3-VL-2B at $\rho=0.2$.
}
\label{tab:ablation}
\scriptsize
\setlength{\tabcolsep}{2.0pt}
\resizebox{\linewidth}{!}{
\begin{tabular}{lccccccccccc}
\toprule
\multirow{2}{*}{\textbf{Variant}}
& \multirow{2}{*}{\textbf{Hallusion\textuparrow}}
& \multirow{2}{*}{\textbf{SQA\textuparrow}}
& \multirow{2}{*}{\textbf{MMBench\textuparrow}}
& \multicolumn{2}{c}{\textbf{MME\textuparrow}}
& \multicolumn{2}{c}{\textbf{MMT\textuparrow}}
& \multirow{2}{*}{\textbf{MathVista\textuparrow}}
& \multicolumn{2}{c}{\textbf{MathVision\textuparrow}}
& \multirow{2}{*}{\textbf{Rel. Avg.\textuparrow}} \\
\cmidrule(lr){5-6}\cmidrule(lr){7-8}\cmidrule(lr){10-11}
& & & 
& \textbf{Perc.} & \textbf{Cog.}
& \textbf{SI} & \textbf{MI}
& 
& \textbf{MINI} & \textbf{Full}
& \\
\midrule
GRACE 
& \textbf{46.8} & \textbf{85.0} & \textbf{73.8} 
& \underline{1512.3} & \underline{682.9} 
& \textbf{58.5} & \textbf{56.3} 
& \underline{54.2} & \textbf{21.7} & \underline{17.2} 
& \textbf{108.8} \\
\midrule
\hspace{0.8em}w/o Hist. 
& \underline{45.9} & \underline{83.3} & 71.0 
& 1502.6 & \textbf{686.8} 
& 56.6 & \underline{54.8} 
& \textbf{55.2} & \underline{17.2} & \textbf{18.4} 
& \underline{105.5} \\
\hspace{0.8em}w/o Ans. 
& 44.5 & 80.8 & 69.7 
& 1495.1 & 655.4 
& 55.2 & 53.3 
& 50.7 & 13.6 & 13.8 
& 97.5 \\
\hspace{0.8em}Target = CoT + Ans 
& \textbf{46.8} & 81.8 & 69.2 
& 1508.1 & 655.0 
& 56.1 & 54.5 
& 50.7 & 15.3 & 14.5 
& 100.2 \\
\hspace{0.8em}Target = Suffix 
& 44.4 & 82.1 & 68.5 
& \textbf{1515.9} & 656.1 
& 55.2 & 52.7 
& 51.3 & 14.4 & 15.1 
& 99.0 \\
\hspace{0.8em}Proxy = Proj. Grad. 
& 45.7 & 82.9 & \underline{71.7} 
& 1501.0 & 661.8 
& \underline{56.7} & \underline{54.8} 
& 49.0 & 12.7 & 14.1 
& 98.3 \\
\bottomrule
\end{tabular}
}
\end{table}

Table~\ref{tab:ablation} reports the ablation results across all benchmarks. 
Both utility components contribute to GRACE. 
Using only answer-oriented alignment remains competitive, reaching $105.5\%$ relative average performance, but still trails full GRACE by $3.3$ points, showing that trajectory consistency provides complementary information beyond answer alignment. 
In contrast, removing the answer-oriented term drops the relative average to $97.5\%$, suggesting that historical consistency alone may favor internally coherent traces that are not necessarily aligned with the final optimization objective.

The target direction also matters. 
Replacing the answer segment with the full trace (\textit{CoT + Ans}) or local suffix reduces the relative average to $100.2\%$ and $99.0\%$, respectively, indicating that the final answer is a more reliable target for step-level utility estimation. 
Finally, replacing our representation-level proxy with projected parameter-space gradients yields only $98.3\%$, confirming that our proxy is not merely an efficiency approximation but also provides a stable signal for reasoning-step valuation.
  
\subsection{Hyperparameter Analysis}
\label{sec:exp_hyper}
 

We study the sensitivity of GRACE to its main hyperparameters on Qwen3-VL-2B at $\rho=0.2$. 
We examine: 
(i) the candidate pool (default $\geq 8$-step subset of MMathCoT-1M vs.\ full MMathCoT-1M);
(ii) the warm-up strategy, including the warm-up ratio $\gamma$ and scoring checkpoints; 
(iii) the history aggregation scheme $\omega_{k,j}$ (uniform / sliding window with size $W$ / EMA with decay $\beta$);
and (iv) the balance coefficient $\alpha\in[0,1]$ for answer-oriented and historical alignment.


\begin{table}[tb]
\centering
\caption{Hyperparameter analysis on Qwen3-VL-2B at $\rho=0.2$.
$\gamma$ denotes the warm-up ratio; $W$ and $\beta$ are used for window and EMA history aggregation, respectively.
For four-checkpoint scoring, $0.25$--$1.0$ denotes checkpoints at $25\%$, $50\%$, $75\%$, and $100\%$ warm-up progress.
}
\label{tab:hyperparameter}
\small
\begin{tabular}{llcccccc}
\toprule
\textbf{Study}
& \textbf{Variant}
& \boldmath$\gamma$
& \textbf{Hist.}
& \boldmath$W$
& \boldmath$\beta$
& \boldmath$\alpha$
& \textbf{Rel. Avg.\textuparrow} \\
\midrule
\multirow{2}{*}{Pool}
& $\geq$8-step
& -- & -- & -- & -- & -- & 100.0 \\
& Full w/o filter
& -- & -- & -- & -- & -- & 99.3 \\
\midrule
\multirow{4}{*}{Warm-up}
& No warm-up
& 0 & uniform & -- & -- & 0.5 & 89.6 \\
& Larger warm-up
& 0.25 & uniform & -- & -- & 0.5 & 99.6 \\
& Four checkpoints
& 0.25--1.0 & uniform & -- & -- & 0.5 & 101.7 \\
& Default
& 0.05 & uniform & -- & -- & 0.5 & 106.6 \\
\midrule
\multirow{3}{*}{History}
& Window
& 0.05 & window & 3 & -- & 0.5 & 104.6 \\
& EMA
& 0.05 & EMA & -- & 0.8 & 0.5 & 105.7 \\
& Uniform
& 0.05 & uniform & -- & -- & 0.5 & 106.6 \\
\midrule
\multirow{4}{*}{Balance}
& $\alpha=0.1$
& 0.05 & uniform & -- & -- & 0.1 & 100.8 \\
& $\alpha=0.5$
& 0.05 & uniform & -- & -- & 0.5 & 106.6 \\
& $\alpha=0.7$ (default)
& 0.05 & uniform & -- & -- & 0.7 & \textbf{108.8} \\
& $\alpha=0.9$
& 0.05 & uniform & -- & -- & 0.9 & \underline{107.7} \\
\bottomrule
\end{tabular}
\end{table}

Table~\ref{tab:hyperparameter} summarizes representative hyperparameter variants.
These results lead to four main observations.
\textbf{Candidate pool.} The $\geq$8-step pool slightly outperforms the unfiltered MMathCoT-1M pool, supporting the use of reasoning-rich traces for stable step-level scoring.
\textbf{Warm-up.} Without warm-up, performance drops to $89.6\%$, while the lightweight default warm-up with $\gamma=0.05$ reaches $106.6\%$, outperforming larger warm-up or multi-checkpoint scoring.
\textbf{History aggregation.} Uniform cumulative averaging performs best among the tested strategies, suggesting that broader reasoning history provides a stable reference direction.
\textbf{Balance coefficient.} $\alpha=0.7$ achieves the best result ($108.8\%$), showing that answer alignment should dominate while still retaining trajectory consistency.
Full per-benchmark results are provided in Appendix~\ref{app:full_hyperparameter}.

\subsection{Computational Cost}
\label{sec:exp_cost}

We focus on the dominant offline cost of feature/signal collection, as the subsequent ranking and top-$\rho$ selection costs are negligible in comparison.
Let $\mathcal{D}_{\mathrm{train}}$ and $\mathcal{D}_{\mathrm{target}}$ denote the candidate pool and the target/validation set, respectively; let $M$ be the number of scoring checkpoints, $K$ the average number of reasoning steps, $d_{\mathrm{proj}}$ the projected-gradient dimension, and $d$ the hidden dimension defined in Sec.~\ref{sec:proxy}. 
We use $C_{\mathrm{fwd}}$ and $C_{\mathrm{bwd}}$ for the cost of one forward and backward pass. 
Table~\ref{tab:cost} summarizes the time and storage complexity of this collection stage.

\begin{table}[tb]
\centering
\caption{Cost complexity of feature/signal collection.}
\label{tab:cost}
\small
\setlength{\tabcolsep}{3pt}
\resizebox{\linewidth}{!}{
\begin{tabular}{lcc}
\toprule
\textbf{Method}
& \textbf{Time Complexity}
& \textbf{Storage Complexity} \\
\midrule
Gradient projection
&
$\mathcal{O}\!\left(
M\bigl(|\mathcal{D}_{\mathrm{train}}|+|\mathcal{D}_{\mathrm{target}}|\bigr)
(C_{\mathrm{fwd}}+C_{\mathrm{bwd}})
\right)$
&
$\mathcal{O}\!\left(
M\bigl(|\mathcal{D}_{\mathrm{train}}|+|\mathcal{D}_{\mathrm{target}}|\bigr)d_{\mathrm{proj}}
\right)$
\\
GRACE gradient proxy
&
$\mathcal{O}\!\left(
M|\mathcal{D}_{\mathrm{train}}|C_{\mathrm{fwd}}
\right)$
&
$\mathcal{O}\!\left(
M|\mathcal{D}_{\mathrm{train}}|(K+1)d
\right)$
\\
\bottomrule
\end{tabular}
}
\end{table}

Gradient-projection methods collect projected gradients for both training and target samples, and each feature requires a forward--backward pass. 
This is expensive because it scales with both $\mathcal{D}_{\mathrm{train}}$ and $\mathcal{D}_{\mathrm{target}}$ and stores $d_{\mathrm{proj}}$-dimensional gradients for all samples.
GRACE simplifies this collection stage. 
It requires no target set or backward computation; instead, it only performs forward passes over candidate training samples. 
Token-level upstream signals are then grouped into $K$ step proxies and one answer proxy in the hidden dimension $d$, enabling step-aware valuation in a single pass per sample. 
In practice, this proxy collection remains lightweight, taking 6.5 single-node hours and 5.7GB of storage on the candidate pool.

Overall, GRACE replaces backward-based gradient extraction with forward-only proxy collection, making step-level reasoning data valuation substantially more efficient while preserving the fine-grained structure needed for our scoring objective.

\section{Related Work}


\paragraph{Reasoning supervision and step-level evaluation.}
Chain-of-thought prompting and supervision improve reasoning by exposing intermediate rationales~\citep{wei2022chain,xu2025llavacot}. 
As reasoning datasets scale, post-training pipelines often curate data using quality indicators, such as final-answer correctness, self-consistency~\citep{wang2023selfconsistency}, reward-model scores, or preference signals~\citep{ouyang2022training,rafailov2023dpo}. 
Process-supervision methods further evaluate intermediate steps with human annotations or verifiers~\citep{mmathcot,uesato2022solving,lightman2023lets,gao2025benchmarking}. 
GRACE instead scores steps with internal optimization signals, without external rewards or step annotations.

\paragraph{Data valuation and efficient post-training.}
Influence functions and TracIn-style estimators measure training-sample effects through optimization dynamics~\citep{koh2017influence,tracin,representerpoint,datashapley}, and scalable variants such as LESS approximate these effects with projected gradients~\citep{datamodels,less}. 
Vision-language data selection methods further explore influence consensus or capability-aware curation~\citep{icons,cadc}, while efficient instruction tuning often relies on diversity~\citep{diverseevol}, difficulty~\citep{wizardlm}, task coverage~\citep{icons}, data quality~\citep{liu2024what,tive,presel,instructmining}, or sample-level influence~\citep{less,icons,cadc}. 
These methods mainly operate at the sample level, whereas GRACE models the ordered internal structure of reasoning traces and aggregates step-level directional utilities for subset selection.

\section{Conclusion}

We present GRACE, a gradient-aligned reasoning data curation method for efficient post-training. Instead of treating a reasoning trace as an indivisible supervision unit, GRACE views it as a sequence of optimization events and evaluates each step through answer-oriented alignment and trajectory consistency. To make this fine-grained valuation scalable, we introduce a representation-level gradient proxy that estimates step-induced update directions from forward-pass upstream signals, avoiding per-step backward computation. Experiments on multimodal reasoning post-training show that GRACE selects compact subsets that match or surpass full-data training, with 20\% curated data reaching 108.8\% of full-data performance and 5\% retaining 100.2\%. These results suggest that the value of reasoning data depends not only on external quality indicators such as correctness, preference, or length, but also on whether its intermediate steps constructively support the optimization trajectory.

\bibliographystyle{unsrtnat} 

\bibliography{ref}


\appendix

\section{Limitations}
\label{app:limitations}

GRACE has several limitations.
First, GRACE estimates data utility from optimization signals observed under a fixed scoring model and warm-up configuration. Although our experiments show consistent gains across selection ratios, hyperparameter variants, and transferred backbones, applying GRACE to substantially different optimizers, training objectives, or model families may require additional empirical validation.
Second, our empirical evaluation focuses on multimodal reasoning post-training with chain-of-thought data. While the results cover mathematical reasoning, general visual question answering, multi-task reasoning, and multi-image reasoning benchmarks, extending GRACE to pure language reasoning, non-CoT instruction data, or reinforcement-learning-based post-training remains an important direction for future work.
Third, while improving data efficiency can reduce post-training cost and make reasoning model development more accessible, more effective curation may also lower the barrier to training stronger models for unintended or harmful downstream uses. This highlights the importance of responsible dataset governance, safety evaluation, and deployment when applying GRACE to capability-enhancing post-training pipelines.

\section{First-order Motivation for Directional Step Utility}
\label{app:directional_step_utility}

For a sample $z$ and a target loss $L^{\mathrm{tar}}(\theta;z)$, consider the step-induced update
\begin{equation}
\theta'=\theta-\eta\nabla_\theta L_k(\theta;z),
\end{equation}
where $\eta>0$. Applying a first-order Taylor expansion at $\theta$ gives
\begin{align}
L^{\mathrm{tar}}(\theta';z)
&=
L^{\mathrm{tar}}(\theta;z)
+
\left\langle
\nabla_\theta L^{\mathrm{tar}}(\theta;z),
\theta'-\theta
\right\rangle
+
O(\|\theta'-\theta\|^2) \\
&=
L^{\mathrm{tar}}(\theta;z)
-
\eta
\left\langle
\nabla_\theta L^{\mathrm{tar}}(\theta;z),
\nabla_\theta L_k(\theta;z)
\right\rangle
+
O(\eta^2).
\end{align}
Therefore,
\begin{equation}
L^{\mathrm{tar}}(\theta';z)-L^{\mathrm{tar}}(\theta;z)
=
-\eta
\left\langle
\nabla_\theta L_k(\theta;z),
\nabla_\theta L^{\mathrm{tar}}(\theta;z)
\right\rangle
+
O(\eta^2).
\end{equation}
Under a small step size, ignoring the higher-order term yields Eq.~\ref{eq:first_order_target}.

The first-order term shows that the local effect of step $s_k$ is governed by the inner product between the step gradient and the target gradient. This inner product can be written as
\begin{equation}
\left\langle
\nabla_\theta L_k,
\nabla_\theta L^{\mathrm{tar}}
\right\rangle
=
\|\nabla_\theta L_k\|
\,
\|\nabla_\theta L^{\mathrm{tar}}\|
\,
A_k^{\mathrm{tar}},
\end{equation}
where
\begin{equation}
A_k^{\mathrm{tar}}
\triangleq
\cos\bigl(
\nabla_\theta L_k,
\nabla_\theta L^{\mathrm{tar}}
\bigr).
\end{equation}
Thus, \(A_k^{\mathrm{tar}}\) corresponds to the normalized directional component of the first-order utility. GRACE uses this normalized form to compare step directions while reducing the effect of gradient scale.

\section{Derivation of Representation-level Gradient Proxy}
\label{app:proxy_derivation}

We derive the representation-level proxy in Sec.~\ref{sec:proxy} from the parameter-space formulation in Sec.~\ref{sec:step}.

Let $\theta_{\mathrm{rep}}$ denote the parameters that produce the final hidden representations. 
For a token-level loss $L_t$, by the chain rule,
\begin{equation}
\nabla_{\theta_{\mathrm{rep}}} L_t
=
\left(\frac{\partial h_t}{\partial \theta_{\mathrm{rep}}}\right)^\top
\frac{\partial L_t}{\partial h_t}
=
J_t^\top u_t,
\label{eq:app_chain}
\end{equation}
where
\[
J_t \triangleq \frac{\partial h_t}{\partial \theta_{\mathrm{rep}}},
\qquad
u_t \triangleq \frac{\partial L_t}{\partial h_t}.
\]

For softmax cross-entropy, let $\ell_t=W_{\mathrm{out}}^\top h_t$,
$p_t=\mathrm{softmax}(\ell_t)$, and $y_t$ be the one-hot label. 
The token-level loss is
\[
L_t=-\sum_{v=1}^{V}(y_t)_v\log p_{t,v},
\qquad
p_{t,v}=\frac{\exp(\ell_{t,v})}{\sum_{r=1}^{V}\exp(\ell_{t,r})}.
\]
The derivative of the softmax probability with respect to the logit is
\[
\frac{\partial p_{t,v}}{\partial \ell_{t,r}}
=
p_{t,v}(\mathbf{1}_{v=r}-p_{t,r}).
\]
Applying the chain rule gives
\begin{align}
\frac{\partial L_t}{\partial \ell_{t,r}}
&=
-\sum_{v=1}^{V}
\frac{(y_t)_v}{p_{t,v}}
\frac{\partial p_{t,v}}{\partial \ell_{t,r}} \nonumber \\
&=
-\sum_{v=1}^{V}
(y_t)_v(\mathbf{1}_{v=r}-p_{t,r})
=
p_{t,r}-(y_t)_r .
\end{align}
Thus, in vector form,
\begin{equation}
\frac{\partial L_t}{\partial \ell_t}
=
p_t-y_t.
\label{eq:app_logit_grad}
\end{equation}


Since the Jacobian of $\ell_t$ with respect to $h_t$ is $W_{\mathrm{out}}^\top$, the upstream gradient at the representation interface is
\begin{equation}
u_t
\triangleq
\frac{\partial L_t}{\partial h_t}
=
\left(\frac{\partial \ell_t}{\partial h_t}\right)^\top
\frac{\partial L_t}{\partial \ell_t}
=
W_{\mathrm{out}}(p_t-y_t).
\label{eq:app_upstream}
\end{equation}

For a token set $\mathcal{T}$, the corresponding representation-parameter gradient is
\begin{equation}
\nabla_{\theta_{\mathrm{rep}}} L(\mathcal{T})
=
\frac{1}{|\mathcal{T}|}
\sum_{t\in\mathcal{T}}J_t^\top u_t.
\label{eq:app_set_grad}
\end{equation}
Thus, $\{u_t\}$ are the common upstream optimization signals that drive updates of the representation-producing parameters through the Jacobian mapping.

Consider two token sets $\mathcal{T}_1$ and $\mathcal{T}_2$. 
Their representation-parameter gradient inner product is
\begin{align}
\left\langle
\nabla_{\theta_{\mathrm{rep}}} L(\mathcal{T}_1),
\nabla_{\theta_{\mathrm{rep}}} L(\mathcal{T}_2)
\right\rangle
&=
\left\langle
\frac{1}{|\mathcal{T}_1|}\sum_{t\in\mathcal{T}_1}J_t^\top u_t,
\frac{1}{|\mathcal{T}_2|}\sum_{t'\in\mathcal{T}_2}J_{t'}^\top u_{t'}
\right\rangle \nonumber \\
&=
\frac{1}{|\mathcal{T}_1||\mathcal{T}_2|}\sum_{t\in\mathcal{T}_1}
\sum_{t'\in\mathcal{T}_2}
u_t^\top
\left(J_tJ_{t'}^\top\right)
u_{t'}.
\label{eq:app_inner}
\end{align}

Accordingly, their exact representation-parameter cosine alignment is
\begin{equation}
\cos\!\left(
\nabla_{\theta_{\mathrm{rep}}} L(\mathcal{T}_1),
\nabla_{\theta_{\mathrm{rep}}} L(\mathcal{T}_2)
\right)
=
\frac{
\sum_{t\in\mathcal{T}_1}
\sum_{t'\in\mathcal{T}_2}
u_t^\top
\left(J_tJ_{t'}^\top\right)
u_{t'}
}{
\left\|
\sum_{t\in\mathcal{T}_1}J_t^\top u_t
\right\|
\left\|
\sum_{t'\in\mathcal{T}_2}J_{t'}^\top u_{t'}
\right\|
},
\label{eq:app_jacobian_cosine}
\end{equation}
where the segment-length normalization factors cancel in the cosine.

Eq.~\ref{eq:app_jacobian_cosine} shows that exact alignment depends on both upstream gradients and Jacobian-induced interactions. 
Exact step-level evaluation of these interactions would require isolating each step loss and explicitly backpropagating it through $\theta_{\mathrm{rep}}$, leading to repeated per-step gradient computation. 
We therefore use an interface-level surrogate that preserves the upstream optimization signals while avoiding explicit construction of the Jacobian-induced geometry, in line with scalable data valuation methods that approximate gradient information in proxy spaces. 
This yields the representation-level proxy
\begin{equation}
g(\mathcal{T})
\triangleq
\frac{1}{|\mathcal{T}|}
\sum_{t\in\mathcal{T}}u_t.
\label{eq:app_proxy}
\end{equation}

For the $k$-th reasoning step and the answer segment, we define
\[
g_k \triangleq g(\mathcal{T}_k),
\qquad
g_{\mathrm{ans}} \triangleq g(\mathcal{T}_{\mathrm{ans}}).
\]
Using this proxy, the gradient-space quantities in Sec.~\ref{sec:step} are instantiated as
\begin{equation}
\widehat{\mathrm{Score}}_k
=
\begin{cases}
\widehat{A}_k^{\mathrm{ans}}, & k=1,\\[2mm]
\alpha \widehat{A}_k^{\mathrm{ans}}
+
(1-\alpha)\widehat{A}_k^{\mathrm{hist}}, & k>1,
\end{cases}
\end{equation}
where
\[
\widehat{A}_k^{\mathrm{ans}}\triangleq\cos(g_k,g_{\mathrm{ans}}),
\quad
\widehat{A}_k^{\mathrm{hist}}\triangleq\cos(g_k,\widehat{r}_k),
\quad
\widehat{r}_k\triangleq
\mathrm{Normalize}\left(\sum_{j<k}\omega_{k,j}g_j\right).
\]

Thus, $\widehat{\mathrm{Score}}_k$ is the computable representation-level approximation of the gradient-space utility $\mathrm{Score}_k$.
The coefficients $\omega_{k,j}$ can instantiate several common history aggregation strategies:
\[
\omega_{k,j} =
\begin{cases}
\displaystyle \frac{1}{k-1}, 
& \text{Uniform}, \\[2mm]
\displaystyle \frac{\mathbf{1}\{\max(1,k-W)\le j<k\}}{\min(W,k-1)}, 
& \text{Sliding window}, \\[2mm]
\displaystyle \frac{\beta^{k-1-j}}{\sum_{r=1}^{k-1}\beta^{k-1-r}}, 
& \text{EMA}.
\end{cases}
\]
where $W\in\mathbb{N}$ denotes the window size, 
$\beta \in [0,1)$ is the decay factor,
and $\mathbf{1}\{\cdot\}$ is the indicator function.

\section{Detailed Algorithm}
\label{app:algorithm}

Algorithm~\ref{alg:grace} provides the complete curation procedure of GRACE, including model warm-up, forward-only proxy extraction, step-level scoring, sample-level aggregation, and top-$\rho$ subset selection.

\begin{algorithm}[t]
\caption{GRACE reasoning data curation}
\label{alg:grace}
\begin{algorithmic}[1]
\REQUIRE Reasoning dataset $\mathcal{D}$, initial model $f_{\theta_0}$, warm-up ratio $\gamma$, selection ratio $\rho$, balance coefficient $\alpha$
\ENSURE Curated subset $\mathcal{S}$

\STATE Warm up $f_{\theta_0}$ on a subset of $\mathcal{D}$ with ratio $\gamma$ to obtain $f_{\theta}$.
\STATE Keep $f_{\theta}$ fixed during scoring and let $W_{\mathrm{out}}$ be its output projection.
\FOR{each sample $z_i=(x_i,\mathbf{s}_i,a_i)\in\mathcal{D}$}
    \STATE Run a forward pass with model $f_\theta$ to obtain token probabilities $p_t$.
    \STATE Compute upstream signals:
    $u_t \gets W_{\mathrm{out}}(p_t-y_t)$,
    where $y_t$ is the ground-truth token.
    \FOR{each reasoning step $s_{i,k}$}
        \STATE $g_{i,k} \gets |\mathcal{T}_{i,k}|^{-1}\sum_{t\in\mathcal{T}_{i,k}}u_t$.
    \ENDFOR
    \STATE $g_{i,\mathrm{ans}} \gets |\mathcal{T}_{i,\mathrm{ans}}|^{-1}\sum_{t\in\mathcal{T}_{i,\mathrm{ans}}}u_t$.
    \FOR{each reasoning step $s_{i,k}$}
        \IF{$k=1$}
            \STATE $\widehat{\mathrm{Score}}_{i,1} \gets \cos(g_{i,1},g_{i,\mathrm{ans}})$.
        \ELSE
            \STATE $\widehat{r}_{i,k} \gets \mathrm{Normalize}\!\left(\sum_{j<k}\omega_{k,j}g_{i,j}\right)$.
            \STATE $\widehat{\mathrm{Score}}_{i,k} \gets
            \alpha\cos(g_{i,k},g_{i,\mathrm{ans}})
            +(1-\alpha)\cos(g_{i,k},\widehat{r}_{i,k})$.
        \ENDIF
    \ENDFOR
    \STATE $V(z_i) \gets K_i^{-1}\sum_{k=1}^{K_i}\widehat{\mathrm{Score}}_{i,k}$.
\ENDFOR
\STATE Select $\mathcal{S}$ as the top $\lceil \rho|\mathcal{D}|\rceil$ samples ranked by $V(z_i)$.
\RETURN $\mathcal{S}$
\end{algorithmic}
\end{algorithm}

\section{Experimental Details}
\label{app:experimental_details}
 
\paragraph{Training data.}
We post-train on MMathCoT-1M~\citep{mmathcot}, a large-scale multimodal mathematical reasoning corpus with chain-of-thought supervision.
Since GRACE evaluates step-level optimization signals, we use reasoning traces with at least eight reasoning steps as the default candidate pool, which provides sufficient granularity for stable step-level utility estimation.
Preliminary experiments show that training on the full MMathCoT-1M does not improve downstream performance over this reasoning-rich pool, suggesting that this filtering does not compromise training effectiveness.
Unless explicitly stated otherwise, all reported full-data baselines and selection ratios are defined with respect to this $\geq$8-step candidate pool.
 
\paragraph{Backbone models.}
Our default backbone is Qwen3-VL-2B-Instruct~\citep{qwen3vl}. 
To examine the transferability of GRACE-selected subsets, we additionally post-train Qwen2.5-VL-3B~\citep{qwen2.5-VL}, LLaVA-1.5-7B~\citep{llava15}, and Qwen3-VL-8B-Instruct on the same curated subset, without re-running data selection.

For data scoring, we warm up the initial model on a $\gamma=0.05$ subset of the candidate pool to obtain the fixed scoring model $f_\theta$, and use a single warm-up checkpoint by default.
The warm-up subset is used only to obtain the fixed scoring model and is not counted as post-training data.
For fairness, all gradient- and proxy-based baselines use the same warm-up checkpoint when applicable.
The default selection ratio is $\rho=0.2$.
The historical reference direction $\widehat{r}_k$ uses uniform aggregation
($\omega_{k,j}=1/(k-1)$, i.e., the historical average), and the balance coefficient is fixed to $\alpha=0.7$.
Post-training and evaluation are conducted with the ms-swift framework~\citep{msswift}.
 
\paragraph{Baselines.}
We compare GRACE against two families of baselines:
(i) \emph{heuristic selectors}---Random (uniform sampling), Longest (longest reasoning traces by token length), and Stepmax (traces with the largest number of steps);
(ii) \emph{state-of-the-art data curation methods}---LESS~\citep{less}, a gradient-projection-based influence method; ICONS~\citep{icons}, a cross-task influence-consensus selector; and CADC~\citep{cadc}, a recent curriculum-aware data curator.
For fair comparison, all baselines select subsets at the same ratio $\rho$ from the same candidate pool and are post-trained with identical training recipes.

\paragraph{Benchmarks.}
We evaluate the post-trained models on a diverse suite of multimodal benchmarks using the VLMEvalKit~\citep{vlmevalkit} backend integrated in ms-swift, covering three categories:
\begin{itemize}
    \item \emph{general visual question answering and perception}---HallusionBench~\citep{hallusionbench}, ScienceQA~\citep{scienceqa}, MMBench~\citep{mmbench}, MME (Perception and Cognition)~\citep{mme};
    \item \emph{multi-task and multi-image reasoning}---MMT-Bench (single-image) and MMT-Bench\_MI (multi-image)~\citep{mmtbench};
    \item \emph{mathematical reasoning}---MathVista~\citep{mathvista}, MathVision\_MINI, and MathVision (full)~\citep{mathvision}.
\end{itemize}
We report task-specific metrics for each benchmark, and additionally report relative average performance (\textit{Rel.\ Avg.}) normalized by the full-data training baseline.
For compactness, tables abbreviate HallusionBench as Hallusion, ScienceQA as SQA, MME Perception/Cognition as Perc./Cog., and MMT-Bench single-/multi-image settings as SI/MI.

Given $B$ evaluation entries, including benchmark sub-scores reported separately, we compute the relative score of a selected subset $\mathcal{S}$ on entry $b$ as
\[
R_b(\mathcal{S}) =
\frac{m_b(\mathcal{S})}{m_b(\mathcal{D}_{\mathrm{full}})} \times 100,
\]
where $m_b(\cdot)$ denotes the metric of entry $b$, and $\mathcal{D}_{\mathrm{full}}$ denotes the full $\geq$8-step candidate pool.
The relative average is then
\[
\mathrm{Rel.\ Avg.} =
\frac{1}{B}\sum_{b=1}^{B} R_b(\mathcal{S}).
\]

\paragraph{Implementation framework.}
All post-training experiments are implemented with ms-swift v3.12.1, following the official recommended SFT recipe for vision-language models. 
We use PyTorch 2.9.0 with CUDA 12.8 and cuDNN 9.10.2, and use DeepSpeed v0.17.6 with ZeRO-2 optimization for distributed training.

\paragraph{Post-training configuration.}
Unless otherwise specified, all models are fine-tuned for one epoch with LoRA. 
We use bfloat16 precision, FlashAttention, padding-free training, sequence packing, and gradient checkpointing. 
LoRA is applied to all linear layers with rank $8$ and scaling factor $32$, while the vision encoder and multimodal aligner are frozen. 
The learning rate is set to $1\times10^{-4}$ with a warm-up ratio of $0.05$. 
The per-device batch size is $1$, the gradient accumulation step is $2$, and the maximum sequence length is $4096$. 
Optimizer and scheduler settings follow the default ms-swift SFT configuration unless otherwise stated.


\paragraph{Hardware.}
Experiments are conducted on a server with eight NVIDIA A800-SXM4-80GB GPUs and approximately 1.0~TiB system memory, running Ubuntu 22.04.5 LTS. 

\begin{table}[tbp]
\centering
\caption{Full hyperparameter results for candidate pool and warm-up strategy on Qwen3-VL-2B at $\rho=0.2$.
For four-checkpoint scoring, $0.25$--$1.0$ denotes checkpoints at $25\%$, $50\%$, $75\%$, and $100\%$ warm-up progress.}
\label{tab:app_hyper_warmup}
\scriptsize
\setlength{\tabcolsep}{2pt}
\resizebox{\linewidth}{!}{
\begin{tabular}{lcccccccccccccc}
\toprule
\multirow{2}{*}{\textbf{Variant}}
& \multirow{2}{*}{\boldmath$\gamma$}
& \multirow{2}{*}{\textbf{Hist.}}
& \multirow{2}{*}{\boldmath$\alpha$}
& \multirow{2}{*}{\textbf{Hallusion}}
& \multirow{2}{*}{\textbf{SQA}}
& \multirow{2}{*}{\textbf{MMBench}}
& \multicolumn{2}{c}{\textbf{MME}}
& \multicolumn{2}{c}{\textbf{MMT}}
& \multirow{2}{*}{\textbf{MathVista}}
& \multicolumn{2}{c}{\textbf{MathVision}}
& \multirow{2}{*}{\textbf{Rel. Avg.}} \\
\cmidrule(lr){8-9}\cmidrule(lr){10-11}\cmidrule(lr){13-14}
& & & & & & 
& \textbf{Perc.} & \textbf{Cog.}
& \textbf{SI} & \textbf{MI}
& 
& \textbf{MINI} & \textbf{Full}
& \\
\midrule
$\geq$8-step pool & -- & -- & -- & 43.7 & 80.2 & 69.3 & 1517.3 & 656.1 & 55.0 & 53.4 & 52.5 & 14.7 & 16.4 & 100.0 \\
Full w/o step filter & -- & -- & -- & 43.2 & 83.4 & 73.5 & 1526.0 & 593.6 & 56.0 & 54.6 & 50.1 & 14.9 & 15.1 & 99.3 \\
\midrule
Warm-up 25\% & 0.25 & uniform & 0.5 & 46.2 & 83.5 & 72.5 & 1506.7 & 622.9 & 56.7 & 54.4 & 50.3 & 15.0 & 13.8 & 99.6 \\
Four checkpoints & 0.25--1.0 & uniform & 0.5 & 45.6 & 85.2 & 72.4 & 1508.5 & 650.0 & 57.2 & 56.2 & 50.7 & 14.8 & 15.8 & 101.7 \\
No warm-up & 0 & uniform & 0.5 & 12.5 & 72.2 & 70.4 & 1486.7 & 465.7 & 54.6 & 53.1 & 51.4 & 15.8 & 16.8 & 89.6 \\
\bottomrule
\end{tabular}
}
\end{table}

\begin{table}[tbp]
\centering
\caption{Full hyperparameter results for history aggregation on Qwen3-VL-2B at $\rho=0.2$.
All variants use $\gamma=0.05$ and $\alpha=0.5$.
\textbf{Bold} and \underline{underlined} values denote the best and second-best \textit{Rel. Avg.}, respectively.}
\label{tab:app_hyper_history}
\scriptsize
\setlength{\tabcolsep}{2pt}
\resizebox{\linewidth}{!}{
\begin{tabular}{llccccccccccccc}
\toprule
\multirow{2}{*}{\textbf{Hist.}}
& \multirow{2}{*}{\textbf{Param.}}
& \multirow{2}{*}{\boldmath$\gamma$}
& \multirow{2}{*}{\boldmath$\alpha$}
& \multirow{2}{*}{\textbf{Hallusion}}
& \multirow{2}{*}{\textbf{SQA}}
& \multirow{2}{*}{\textbf{MMBench}}
& \multicolumn{2}{c}{\textbf{MME}}
& \multicolumn{2}{c}{\textbf{MMT}}
& \multirow{2}{*}{\textbf{MathVista}}
& \multicolumn{2}{c}{\textbf{MathVision}}
& \multirow{2}{*}{\textbf{Rel. Avg.}} \\
\cmidrule(lr){8-9}\cmidrule(lr){10-11}\cmidrule(lr){13-14}
& & & & & &
& \textbf{Perc.} & \textbf{Cog.}
& \textbf{SI} & \textbf{MI}
&
& \textbf{MINI} & \textbf{Full}
& \\
\midrule
\multirow{6}{*}{Window}
& $W=2$ & 0.05 & 0.5 & 46.1 & 83.0 & 72.7 & 1508.8 & 665.0 & 59.3 & 57.0 & 52.9 & 14.5 & 17.2 & 103.5 \\
& $W=3$ & 0.05 & 0.5 & 45.8 & 84.6 & 72.8 & 1504.5 & 692.1 & 57.2 & 55.8 & 52.0 & 17.1 & 16.7 & 104.6 \\
& $W=4$ & 0.05 & 0.5 & 44.5 & 83.7 & 73.3 & 1504.0 & 610.4 & 57.0 & 54.9 & 54.7 & 17.8 & 16.5 & 103.6 \\
& $W=5$ & 0.05 & 0.5 & 44.3 & 85.8 & 74.4 & 1494.6 & 577.5 & 57.7 & 56.5 & 53.3 & 16.4 & 16.8 & 103.0 \\
& $W=6$ & 0.05 & 0.5 & 45.1 & 82.8 & 71.4 & 1501.8 & 693.9 & 57.3 & 55.3 & 44.2 & 17.4 & 15.8 & 102.0 \\
& $W=8$ & 0.05 & 0.5 & 47.5 & 84.9 & 73.3 & 1486.8 & 652.9 & 57.5 & 55.0 & 53.8 & 17.1 & 16.1 & 104.2 \\
\midrule
\multirow{7}{*}{EMA}
& $\beta=0.70$ & 0.05 & 0.5 & 46.4 & 84.1 & 73.3 & 1505.0 & 671.8 & 56.7 & 55.9 & 51.1 & 18.4 & 16.3 & 104.7 \\
& $\beta=0.75$ & 0.05 & 0.5 & 45.6 & 84.7 & 73.5 & 1506.2 & 658.6 & 57.3 & 55.8 & 49.8 & 19.4 & 16.1 & 104.7 \\
& $\beta=0.80$ & 0.05 & 0.5 & 46.6 & 84.3 & 72.7 & 1496.4 & 669.6 & 57.7 & 55.5 & 51.0 & 20.1 & 16.4 & \underline{105.7} \\
& $\beta=0.85$ & 0.05 & 0.5 & 46.0 & 85.5 & 74.3 & 1493.9 & 588.9 & 57.2 & 55.8 & 52.9 & 18.8 & 16.1 & 104.1 \\
& $\beta=0.90$ & 0.05 & 0.5 & 46.1 & 83.8 & 72.7 & 1508.2 & 666.1 & 57.1 & 54.5 & 54.1 & 17.1 & 16.4 & 104.1 \\
& $\beta=0.95$ & 0.05 & 0.5 & 45.9 & 84.5 & 72.4 & 1498.1 & 651.1 & 57.1 & 56.0 & 53.8 & 17.4 & 15.3 & 103.4 \\
& $\beta=0.99$ & 0.05 & 0.5 & 45.9 & 85.5 & 73.1 & 1483.4 & 610.7 & 57.3 & 55.8 & 51.7 & 15.8 & 15.5 & 101.7 \\
\midrule
Uniform
& -- & 0.05 & 0.5 & 47.3 & 85.5 & 75.1 & 1509.6 & 678.2 & 57.5 & 56.1 & 52.2 & 19.4 & 16.5 & \textbf{106.6} \\
\bottomrule
\end{tabular}
}
\end{table}

\begin{table}[tbp]
\centering
\caption{Full hyperparameter results for the balance coefficient $\alpha$ on Qwen3-VL-2B at $\rho=0.2$.
All variants use $\gamma=0.05$.
\textbf{Bold} and \underline{underlined} values denote the best and second-best \textit{Rel. Avg.}, respectively.}
\label{tab:app_hyper_alpha}
\scriptsize
\setlength{\tabcolsep}{2pt}
\resizebox{\linewidth}{!}{
\begin{tabular}{llcccccccccccc}
\toprule
\multirow{2}{*}{\textbf{Hist.}}
& \multirow{2}{*}{\textbf{Param.}}
& \multirow{2}{*}{\boldmath$\alpha$}
& \multirow{2}{*}{\textbf{Hallusion}}
& \multirow{2}{*}{\textbf{SQA}}
& \multirow{2}{*}{\textbf{MMBench}}
& \multicolumn{2}{c}{\textbf{MME}}
& \multicolumn{2}{c}{\textbf{MMT}}
& \multirow{2}{*}{\textbf{MathVista}}
& \multicolumn{2}{c}{\textbf{MathVision}}
& \multirow{2}{*}{\textbf{Rel. Avg.}} \\
\cmidrule(lr){7-8}\cmidrule(lr){9-10}\cmidrule(lr){12-13}
& & & & & 
& \textbf{Perc.} & \textbf{Cog.}
& \textbf{SI} & \textbf{MI}
&
& \textbf{MINI} & \textbf{Full}
& \\
\midrule
\multirow{9}{*}{Uniform}
& \multirow{9}{*}{--}
& 0.9 & 46.6 & 84.7 & 72.4 & 1506.0 & 683.9 & 56.3 & 54.2 & 52.9 & 21.4 & 17.9 & \underline{107.7} \\
& & 0.8 & 46.2 & 82.5 & 71.5 & 1496.1 & 678.2 & 56.3 & 54.0 & 53.4 & 18.4 & 17.1 & 104.7 \\
& & 0.7 & 46.8 & 85.0 & 73.8 & 1512.3 & 682.9 & 58.5 & 56.3 & 54.2 & 21.7 & 17.2 & \textbf{108.8} \\
& & 0.6 & 45.7 & 83.5 & 71.1 & 1501.3 & 682.1 & 56.9 & 54.1 & 52.7 & 16.4 & 16.5 & 103.2 \\
& & 0.5 & 47.3 & 85.5 & 75.1 & 1509.6 & 678.2 & 57.5 & 56.1 & 52.2 & 19.4 & 16.5 & 106.6 \\
& & 0.4 & 44.0 & 81.6 & 70.0 & 1494.5 & 686.1 & 54.7 & 53.2 & 52.5 & 17.1 & 15.7 & 101.7 \\
& & 0.3 & 47.1 & 82.7 & 70.0 & 1496.6 & 667.9 & 55.2 & 53.8 & 51.1 & 18.4 & 16.5 & 103.6 \\
& & 0.2 & 45.7 & 85.1 & 70.8 & 1500.8 & 665.0 & 56.9 & 54.8 & 50.3 & 17.1 & 16.0 & 102.8 \\
& & 0.1 & 43.9 & 83.8 & 71.4 & 1516.0 & 647.5 & 56.6 & 55.1 & 53.6 & 13.8 & 15.9 & 100.8 \\
\midrule
\multirow{9}{*}{Window}
& \multirow{9}{*}{$W=3$}
& 0.9 & 46.3 & 83.7 & 71.5 & 1493.9 & 686.4 & 57.0 & 54.7 & 54.6 & 21.7 & 16.6 & 107.2 \\
& & 0.8 & 45.1 & 81.9 & 71.8 & 1510.2 & 677.5 & 57.2 & 54.7 & 52.5 & 18.4 & 17.0 & 104.6 \\
& & 0.7 & 44.5 & 84.0 & 72.5 & 1496.6 & 673.6 & 56.8 & 54.3 & 52.3 & 17.8 & 16.4 & 103.7 \\
& & 0.6 & 44.7 & 82.6 & 71.1 & 1503.7 & 695.0 & 56.3 & 54.2 & 52.1 & 15.5 & 16.2 & 102.0 \\
& & 0.5 & 45.8 & 84.6 & 72.8 & 1504.5 & 692.1 & 57.2 & 55.8 & 52.0 & 17.1 & 16.7 & 104.6 \\
& & 0.4 & 46.1 & 83.7 & 71.1 & 1496.1 & 690.0 & 57.1 & 54.7 & 53.1 & 17.4 & 15.6 & 103.6 \\
& & 0.3 & 46.4 & 83.7 & 71.8 & 1505.3 & 670.0 & 56.4 & 53.9 & 53.0 & 18.8 & 15.5 & 104.0 \\
& & 0.2 & 44.1 & 83.1 & 72.0 & 1513.2 & 677.1 & 56.1 & 53.7 & 52.9 & 17.8 & 15.5 & 102.8 \\
& & 0.1 & 45.6 & 83.6 & 71.6 & 1503.0 & 651.8 & 56.0 & 55.2 & 51.7 & 18.4 & 14.7 & 102.6 \\
\midrule
\multirow{9}{*}{EMA}
& \multirow{9}{*}{$\beta=0.8$}
& 0.9 & 46.1 & 83.6 & 72.5 & 1491.7 & 661.8 & 57.2 & 54.7 & 53.5 & 17.4 & 17.0 & 104.4 \\
& & 0.8 & 45.2 & 82.9 & 70.2 & 1498.7 & 672.5 & 56.7 & 54.5 & 51.7 & 16.4 & 17.2 & 103.0 \\
& & 0.7 & 44.1 & 83.7 & 71.7 & 1511.6 & 689.3 & 56.2 & 54.7 & 52.4 & 15.8 & 16.7 & 102.8 \\
& & 0.6 & 47.4 & 83.9 & 70.5 & 1501.0 & 692.9 & 55.3 & 53.7 & 53.6 & 18.1 & 17.4 & 105.1 \\
& & 0.5 & 46.6 & 84.3 & 72.7 & 1496.4 & 669.6 & 57.7 & 55.5 & 51.0 & 20.1 & 16.4 & 105.7 \\
& & 0.4 & 45.5 & 82.7 & 71.6 & 1498.3 & 677.9 & 56.4 & 54.6 & 53.1 & 17.4 & 16.9 & 104.0 \\
& & 0.3 & 45.0 & 84.5 & 71.4 & 1506.5 & 659.6 & 55.8 & 53.9 & 52.5 & 17.4 & 16.8 & 103.4 \\
& & 0.2 & 44.9 & 85.1 & 72.3 & 1512.8 & 635.4 & 56.4 & 54.5 & 54.2 & 16.1 & 16.3 & 102.7 \\
& & 0.1 & 43.9 & 82.4 & 69.7 & 1502.4 & 656.1 & 55.9 & 53.7 & 53.6 & 16.4 & 16.7 & 102.1 \\
\bottomrule
\end{tabular}
}
\end{table}

\section{Full Hyperparameter Results}
\label{app:full_hyperparameter}

We provide the full per-benchmark hyperparameter results in Tables~\ref{tab:app_hyper_warmup}--\ref{tab:app_hyper_alpha}.



\end{document}